\documentclass[letterpaper, 10pt, conference]{ieeeconf}
\IEEEoverridecommandlockouts
\usepackage{cite}
\usepackage{amsmath,amssymb,amsfonts}
\usepackage{algorithm}
\usepackage{algpseudocode}
\usepackage{graphicx}
\usepackage{textcomp}
\usepackage{xcolor}
\usepackage{subfigure}
\def\BibTeX{{\rm B\kern-.05em{\sc i\kern-.025em b}\kern-.08em
    T\kern-.1667em\lower.7ex\hbox{E}\kern-.125emX}}
\usepackage{theorem}

\usepackage{geometry}
\begin{document}

\title{
\huge 
\vspace{9pt}
Confidence-rich Localization and Mapping based on Particle Filter for Robotic Exploration}
\author{Yang Xu, \IEEEmembership{Student Member, IEEE,} Ronghao Zheng$^{\dagger}$, \IEEEmembership{Member, IEEE,}
        Senlin Zhang, \IEEEmembership{Member, IEEE,}\\
        and Meiqin Liu, \IEEEmembership{Senior Member, IEEE}
\thanks{$^1$Yang Xu, Ronghao Zheng and Senlin Zhang are with the College of Electrical Engineering, Zhejiang University, Hangzhou 310027, China. 
		\texttt{\{xuyang94,rzheng,slzhang\}@zju.edu.cn}}
\thanks{$^2$Meiqin Liu is with the Institute of Artificial Intelligence and Robotics, Xi'an Jiaotong University, Xi'an 710049, China. \texttt{liumeiqin@zju.edu.cn}}
\thanks{$^3$All authors are also with the State Key Laboratory of Industrial Control Technology, Zhejiang University, Hangzhou 310027, China.}
\thanks{$^{\dagger}$Corresponding author}
}

\maketitle
\begin{abstract}
This paper mainly studies the localization and mapping of range sensing robots in the confidence-rich map (CRM) and then extends it to provide a full state estimate for information-theoretic exploration.  
Most previous works about active simultaneous localization and mapping and exploration always assumed the known robot poses or utilized inaccurate information metrics to approximate pose uncertainty, resulting in imbalanced exploration performance and efficiency in the unknown environment. This inspires us to extend the confidence-rich mutual information (CRMI) with measurable pose uncertainty.
Specifically, we propose a Rao-Blackwellized particle filter-based localization and mapping scheme (RBPF-CLAM) for CRM, then we develop a new closed-form weighting method to improve the localization accuracy without scan matching. We further derive the uncertain CRMI (UCRMI) with the weighted particles by a more accurate approximation.
Simulations and experimental evaluations show the localization accuracy and exploration performance of the proposed methods.
\end{abstract}

%

\section{Introduction}
\begin{figure}
	\centering
	\subfigure[]{
		\begin{minipage}[t]{0.5\linewidth}
			\centering
			\includegraphics[width=1.8in]{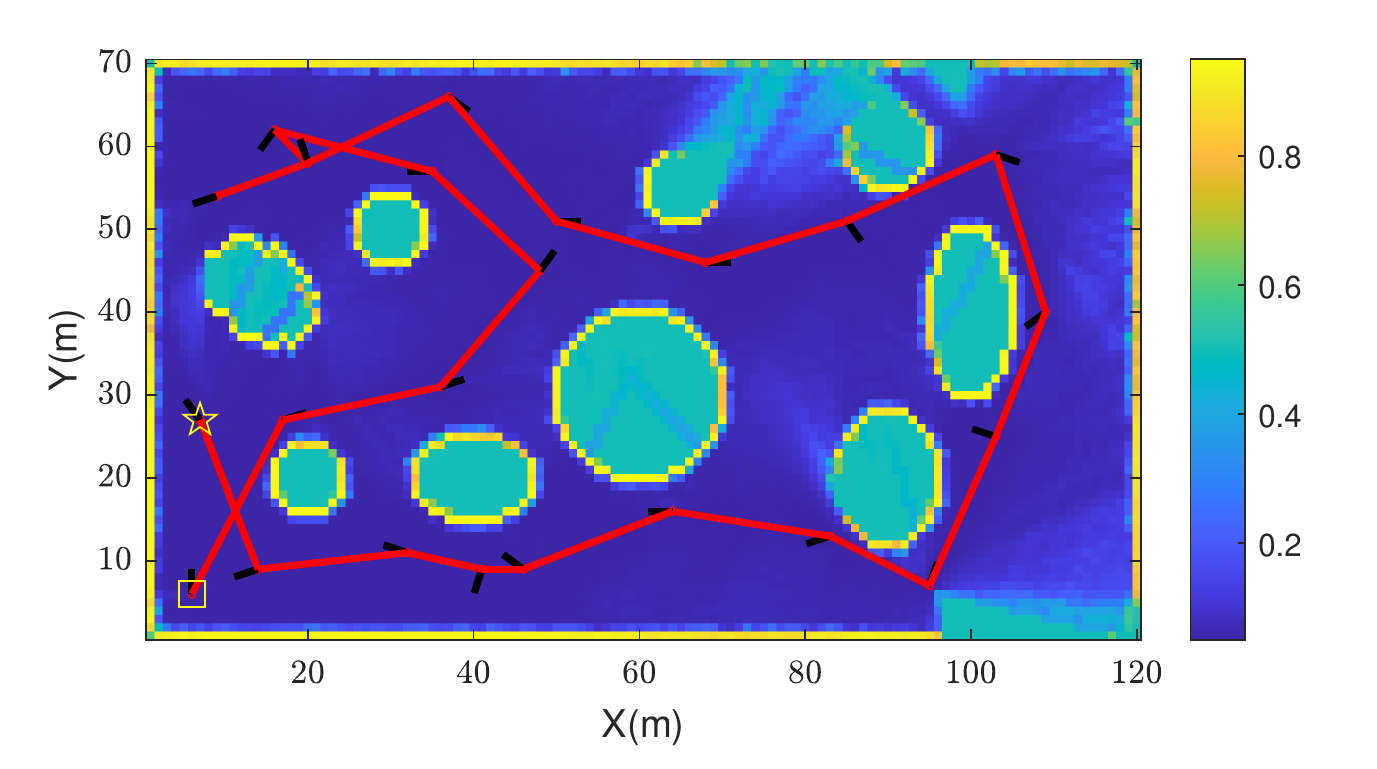}
		\end{minipage}%
	}%
	\subfigure[]{
		\begin{minipage}[t]{0.5\linewidth}
			\centering
			\includegraphics[width=1.8in]{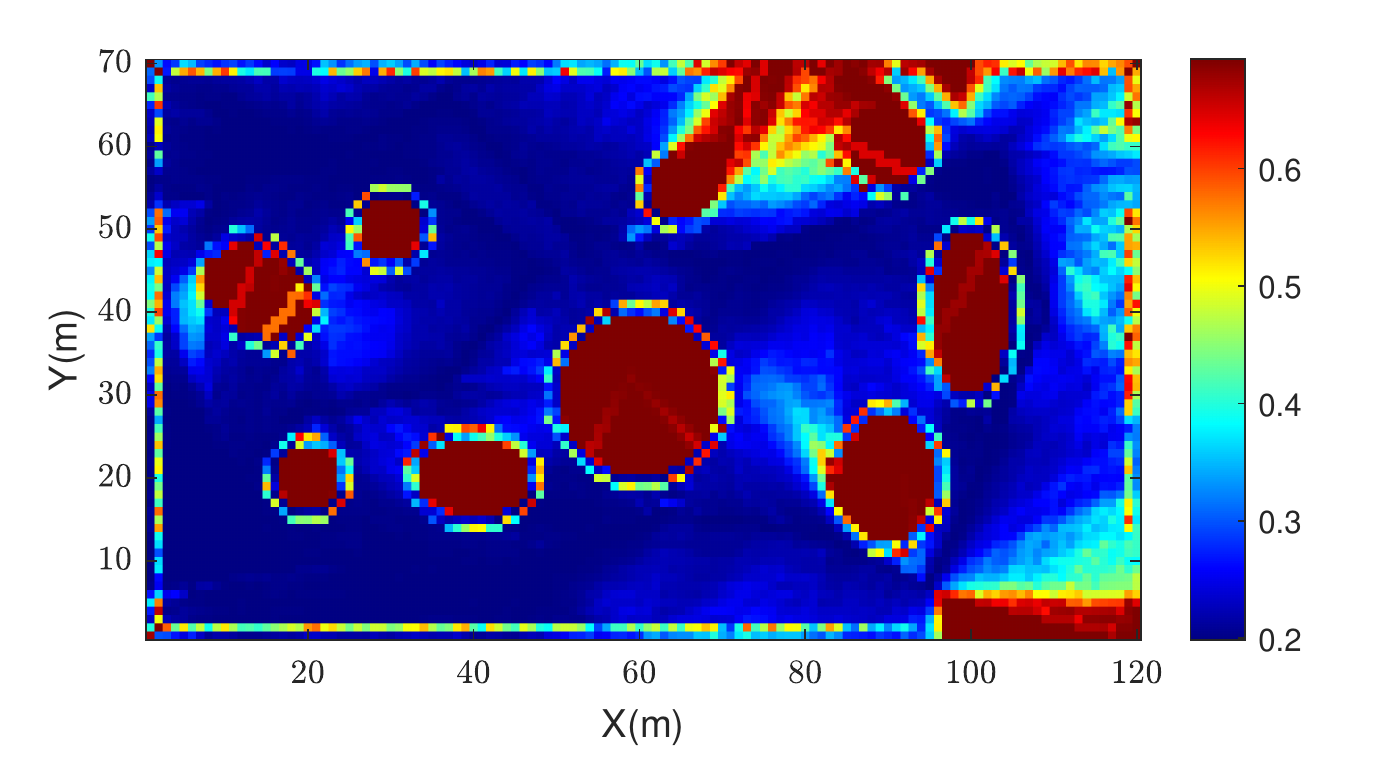}
		\end{minipage}%
	}%
	\centering
	\caption{CRMI-based active robot exploration in an unknown unstructured environment. (a) Informative trajectory generated greedily and the resulting confidence-rich map, (b) Resulting CRMI surface. Note that the yellow square and star are the start and end point, respectively. A minimum information threshold is set to select more informative exploration actions given a budget, e.g. the lower right and top right areas are less informative than the threshold and thus unexplored. Note that the scale of MI is in [0,1] bits in this paper.}
	\label{active}
\end{figure}

Robot exploration has been more prevalent in information gathering tasks such as environment monitoring, objective search and rescue, etc \cite{bircher2016receding,papachristos2017uncertainty,schmid2020efficient,cieslewski2017rapid}. 
Among these methods, information-based robot exploration methods mainly use information-theoretic metrics, such as Shannon's entropy \cite{stachniss2005information, amigoni2010information, carrillo2018autonomous}, mutual information (MI) \cite{julian2014mutual, jadidi2018gaussian, yang2021crmi}, to evaluate the expected utility of candidate actions and decide where to construct the map in the next step or even further. 
A typical exploration example is in Fig.~\ref{active}.

Specially, reducing the pose uncertainty could significantly minimize the mapping uncertainty at the same time, e.g., an information-based controller considering the pose uncertainty reduction can drive the robot to a place where it is more likely to find a loop closure, thereby optimizing the whole trajectory accuracy and the resulting map substantially.

To handle the pose uncertainty, Bourgault \textit{et al.} \cite{bourgault2002information} introduced information metrics into the robot exploration task to choose control policies adaptively, maximizing the map information and minimizing the pose uncertainty under a feature-based extended Kalman filter simultaneous localization and mapping (SLAM) framework. Stachniss \textit{et al.} \cite{stachniss2005information} studied further and made no assumptions about distinguishable landmarks when mapping an occupancy grid using raw laser data. They also utilized a Rao-Blackwellized particle filter representing the maps and poses to evaluate the expected information gain of an action.
Valencia \textit{et al.} \cite{valencia2012active} and Popovi{\'c} \textit{et al.} \cite{popovic2020informative} used a similar way to compute the trajectory entropy, i.e.,  using the average over the uncertainty of different poses along the path, but made the assumption of a multivariate Gaussian distribution over the path in the Pose SLAM context. Similarly, \cite{vallve2015active} used Pose SLAM to estimate the pose and generated candidate paths using RRT* planner. 

\subsection{Related Work}
Most robotic exploration techniques prefer to use the occupancy grid (OG) mapping because of the efficiency of querying and update, as well as the convenience to localize. 
A representative example was the OG map-based MI (OGMI). In \cite{julian2014mutual}, Julian \textit{et al.} computed Shannon's mutual information between new observation and OG maps (OGMs) at candidate poses, and rigorously proved OGMI-based reward function will guide the robot move to the unexplored space. 

Differ from the above methods based on traditional OGMs, Jadidi \textit{et al.} \cite{jadidi2014exploration,jadidi2018gaussian} developed the MI based on Gaussian process (GP) maps. In contrast, GP maps can utilize different kernel methods to train the sparse data sampled from sensor observations and learn a continuous occupancy map (COM). This data-driven mapping scheme considers the inherent correlations between map points and allows arbitrary resolution, and extrapolates to infer the points in unknown areas. \cite{jadidi2017warped} extended the GP mapping with warped GPs using modified kernels for the consideration of pose uncertainty and its propagation into map inference. This consideration has also been introduced to the sampling-based motion planners aimed at robotic exploration, developing two information functions for information gathering, i.e. GP variance reduction (GPVR) and uncertain GPVR (UGPVR) \cite{ghaffari2019sampling}.

Nevertheless, the GPVR, as well as UGPVR, still suffer expensive computational costs, i.e., $O(N_t^3)$ ($N_t$: number of training points) from the learning-based GP mapping due to the inevitable inversion operations of the covariance matrix, which may impede its online performance running on inexpensive robotic platforms. 

More recently, confidence-rich mutual information (CRMI) \cite{yang2021crmi} has been proposed based on the confidence-rich map (CRM) \cite{agha2019confidence} to offer a more accurate information metric for exploration. It captures the continuous dense belief distribution over map occupancy and allows the efficient update, which balances the efficiency of OGMI and the accuracy of GPVR/UGPVR.

However, CRMI, as well as OGMI, focuses more on minimizing the mapping uncertainty and less on the pose uncertainty by assuming the robot's poses are always known during the exploration. 
Though the localization can be conducted as an independent method such as the feature-based SLAM \cite{mourikis2007multi}, it may not work in the more general environmental settings of unstructured scenes that lack easily extractable features, such as planetary exploration and underwater tasks\cite{yang2022robust}. 
Thus, the pose uncertainty in exploration needs to be considered explicitly and a trade-off has to be made between tractability and accuracy.

\subsection{Motivations and Contributions}

Inspired by \cite{ghaffari2019sampling} and \cite{agha2019confidence}, and on the basis of our previous work \cite{yang2021crmi}, this paper studies further based on CRMI and takes the pose uncertainty into account to balance the expected information gain of mapping and pose uncertainty reduction in the evaluation of candidate control policies. We further derive and approximate the expected information gain considering pose uncertainty based on the weighted particles in a more accurate way.

The contributions of this paper mainly are as follows:

1) We propose a Monte Carlo localization scheme named RBPF-CLAM for the confidence-rich map based on Rao-Blackwellized particle filter (RBPF), allowing implementation in unstructured and featureless environments;

2) To improve the localization accuracy, we further derive a closed-form importance weighting approach using the continuous map belief, without a prior map or relying on scan matching results;

3) Moreover, we use the weighted particles of the RBPF-CLAM to explicitly define and approximate the pose uncertainty, and then combine it with CRMI as a new information metric named \textit{uncertain CRMI} (UCRMI), which performs better than other modern information functions.

The remaining paper is organized as follows. Section~II presents the relevant preliminaries and Section~III gives the improved particle filter-based localization methods for CRMs. Section~IV describes the proposed UCRMI considering pose uncertainty. Experimental results and discussions are given in Section~V. We conclude this paper in Section~VI.

\section{Preliminaries}

\subsection{Confidence-rich Grid Map}
In this paper, we mainly consider the widely used beam-based range sensors with finite sensing range, such as LiDAR and sonar.
Here are several related definitions and assumptions. 

Consider a static 2D occupancy grid comprised of $n$ grid cells as a random variable $M=\{M^{[1]},M^{[2]},\dots,M^{[n]}\}$. The occupancy values $m=\{m^{[1]},\dots,m^{[i]},\dots,m^{[n]}\}(m^{[i]} \in [0,1])$ over the set of grid cells of $M$ are defined to denote the continuous occupancy level of a cell. The observation and the robot's state at time step $k$ is modeled as a random variable $Z_k$ taking value $z_k$ and a pose vector $x_k$, respectively.

CRMs relax the binary occupancy assumption and keep the independent grid assumption in OGMs, but take the measurement dependencies between map cells in a sensor cone into account by introducing the beam-based sensor cause model (SCM) \cite{agha2019confidence}. CRMs encode the dependencies into a joint probability distribution over the grid cells, i.e. the map belief. Specifically, the map belief $b^m_k:=p(m|z_{1:k},x_{1:k})$ is defined by the sensor observations $z_{1:k}$ and robot poses $x_{1:k}$, which can be marginalized onto each cell as dense representation: 
\begin{equation*}
b^m_k =
\{b^{m^{[1]}}_k,\dots,b^{m^{[i]}}_k,\dots,b^{m^{[n]}}_k\},~b^{m^{[i]}}_k=p({m^{[i]}}|z_{1:k},x_{1:k}).
\end{equation*}

Thus the high-dimensional belief can be stored via each marginal cells' belief $b^{m^{[i]}}_{k}$ and updated recursively with the mapping scheme $\tau$ \cite{agha2019confidence}:
\begin{equation}
b^{m^{[i]}}_{k} = \tau^{m^{[i]}}( b^{m^{[i]}}_{k-1}, z_k, x_k).
\end{equation}

Note that the robot poses are generally assumed to be known when constructing the CRMs. 

\subsection{CRMI: MI based on CRMs}
Information-theoretic exploration approaches with known poses aim to find a future pose maximizing the MI between the grid map $M$ and new observation $Z_{k}$. Generally, MI can be defined as the functionals of probability distributions by using cross-entropy and Shannon's entropy. 
Julian \textit{et al.} \cite{julian2014mutual} proposed Shannon MI based on the OGM and rigorously proved the attractive behavior to unexplored space for robots driven by information-based controllers. 
As a more expressive MI metric, CRMI considers the non-parametric continuous belief distribution over map occupancy for each cell, which is more descriptive than the MI derived from OGM and the underlying Bernoulli distribution, consequently. 

In particular, in an unknown environment, the CRMI under the special assumption of known robot poses can be defined as follows\cite{yang2021crmi}:
\begin{align}\label{eq:crmi}
&I_m(M;Z_{k}=z_k|z_{1:k-1},x_{1:k-1}) \nonumber \\ 
=&h(M|z_{1:k-1},x_{1:k-1}) - h(M|Z_{k}=z_k,z_{1:k-1},x_{1:k-1}) \nonumber \\ 
=&\int_{z_k\in {\mathcal Z}}p(z_k|z_{1:k-1},x_{1:k-1})\int_{m\in {\mathcal M}} b_{k}^m\log b_{k}^m dmdz  \nonumber \\
-&\int_{m\in{\mathcal M}}b_{k-1}^{m}\log b_{k-1}^{m} dm
\end{align}
where $\mathcal Z$ is the measurement range, $\mathcal M \in [0,1]$ is the occupancy level. Our example of CRMI is open-sourced here\footnote{https://github.com/Shepherd-Gregory/CRMI}.

\section{Monte Carlo Localization for Confidence-rich Maps}
In this section, we present the Monte Carlo-based localization for general CRMs constructed by range sensors.
\subsection{Confidence-rich Mapping using Particle Filter}

Consider a classic full SLAM problem that could be factorized via the well-known Rao-Blackwellization:
\begin{align}
    &p(x_{1:k},m|z_{1:k},u_{0:k-1}) \nonumber \\ =&p(m|x_{1:k},z_{1:k})  \cdot p(x_{1:k}|z_{1:k},u_{0:k-1})={b_k^m} \cdot bel_k(x) \label{eq:rb}
\end{align}
where $bel_k(x):=p(x_{1:k}|z_{1:k},u_{0:k-1})$ is defined as the posterior over robot trajectories, $u_{0:k-1}$ are the past odometry measurements.

As in Eq.~\eqref{eq:rb}, RBPF utilizes a set of particles to approximate the distribution over trajectories and each particle generates one trajectory hypothesis. We have already introduced how to calculate the posterior distribution $b^m$ over map occupancy and solve the mapping problem with known poses in III.~A. To compute the pose posterior $bel_k(x)$, in the context of SLAM using range sensors, the particle filter would be more appropriate.

Grisetti \textit{et al.} \cite{grisetti2007improved} introduced an efficient RBPF-based grid mapping scheme based on sampling importance resampling (SIR). The particle set $S_k=\{s_k^j;j=1,\dots,n_p\}$ contains $n_p$ particles, where the robot pose $x_k^j$ and its corresponding importance weight $\omega_k^j$ are stored in each particle $s_k^j$. The main steps of SIR for CRM are as follows:

1) \textit{State Prediction}: The probabilistic odometry model $p(x_k|x_{k-1},u_k)$ is applied for the partially observable system. This model can also be used as the proposal distribution $\pi$ to sample the particles of next step.

2) \textit{Update the weights}: The pose belief $bel_k(x)$ can be updated recursively using the particle set representing the previous pose belief:
\begin{multline}
    bel_k(x)= \xi \cdot \\
    \begin{matrix} \underbrace{ p(x_{1:k}|z_{1:k-1},u_{0:k-1}) } \\ :=\overline{bel}_k(x) \end{matrix}
    \cdot \begin{matrix} \underbrace{ p(z_k|x_{1:k},z_{1:k-1},u_{0:k-1}) } \\ :=\omega \end{matrix}
\end{multline}
where $\xi$ is the normalization constant. $\overline{bel}_k(x)$ is the current pose belief not corrected by new observation, and the measurement likelihood represents the particles' weights $\omega \propto  bel_k(x)/\overline{bel}_k(x)$.

A classic method to get the expected pose from the candidate particles is to conduct a scan matching algorithm and then obtain a Gaussian proposal distribution $\pi \sim \mathcal N(\mu_k^j,\Sigma_k^j)$, where $\mu_k^j$ and $\Sigma_k^j$ are the mean and variance respectively computed by the Monte Carlo simulation, $\eta_k^j$ is the normalization factor. Hence the weights can be updated by: $\omega_k^j = \omega_{k-1}^j \cdot \eta_k^j$.

If the scan matching fails, an alternative approach to update the particle's importance weight is to assume a fixed map constructed in the latest mapping process and to combine it with the odometry motion model:
\begin{equation}
    \omega_k^j \approx \omega_{k-1}^j \cdot p(z_k|x_k^j,\hat{m}^j_{k-1}),
\end{equation}
where the map of real-valued occupancy can be derived by taking the mathematical expectation of the map belief $b^m$:
\begin{equation}
    \hat{m}:=\mathbb{E}[m]=\int{mb^m}dm, m\in[0,1].
\end{equation}

3) \textit{Resampling the particles}: According to their weights, the particles will be drawn to replace the old ones, which could approximate the continuous distribution by using finite particles. The number of effective particles is defined by the normalized weights: $N_{eff} = 1/\sum_1^{n_p}(\bar \omega^i_k)^2$. This step need to be done when $N_{eff}<n_p/2$. The new weights will be the same after resampling.

4) \textit{Map construction}: Using the CRM mapping scheme, the associated map belief $\{b_k^m\}^j$ of each particle $s_k^j$ can be updated based on the past trajectory $x_{1:k}^j$ and measurements $z_{1:k}$.

\subsection{Improved Weighting using Closed-form CRM}
In the importance weights update procedure, according to \cite{grisetti2007improved}, the proposal distribution is suboptimal, especially when the on-board sensor measurements are more accurate than the odometry estimates.
Instead, we can take the advantage of the continuous map belief in CRM to improve the weight computation and the localization accuracy. 

In a Bayesian framework, we explicitly incorporate the map into the measurement likelihood function:
\begin{align}
    \label{eq:belx}
    &~bel_k(x) \nonumber\\
    &=\xi \overline{bel}_k(x) \int_m p(z_k|m,x_{1:k},z_{1:k-1})p(m|x_{1:k-1},z_{1:k-1})dm \nonumber\\
    &=\xi \overline{bel}_k(x) \int_m p(z_k|m,x_{1:k},z_{1:k-1})b^m_{k-1} dm.
\end{align}

According to the definition of posterior map belief, $b^m_{k-1}$ is a sufficient statistic for all previous poses $x_{1:k-1}$ and observations $z_{1:k-1}$, the following expression holds: 
\begin{equation}
    p(z_k|m,x_{1:k},z_{1:k-1}) \approxeq p(z_k|m,x_k,b^m_{k-1}).
\end{equation}

Similarly, the belief $b_{k}^m$ is also a sufficient statistic for the current candidate pose $x_k$ and the previous map belief $b_{k-1}^m$, so we can get: 
\begin{equation}
    p(z_k|m,x_k,b^m_{k-1}) \approxeq p(z_k|b_{k}^m).
\end{equation}

Thus, Eq.~\eqref{eq:belx} can be rewritten as:
\begin{equation}
    bel_k(x) \propto \overline{bel}_k(x) \int_m p(z_k|b_{k}^m) b^m_{k-1} dm.
\end{equation}

Now the importance weight of particle $j$ storing the pose can be defined as:
\begin{equation}
    \label{eq:wei}
    w^j_k:=\int_m p(z_k|\{b_{k}^m\}^j) \{b_{k-1}^m\}^j dm.
\end{equation}

Here we can compute this weight in a closed-form way similar to \cite{yang2021crmi}. For a beam-based range sensor, under the assumption of independent sensor beams, we can decompose the current measurement/scan $z_k$ into $n_z$ independent beams, then compute the measurement likelihood on each beam $z_k^{[l]}, l\in \{1,\dots, n_z\}$, hence Eq.~\eqref{eq:wei} can be approximated by multiplying the individual likelihood on each beam. 

Particularly, the beam-based likelihood can be derived from the sensor cause model (See Eq.~(8) in \cite{yang2021crmi}). Consequently, the improved weight can be computed as follows:
\begin{equation}
    \label{eq:sumw}
    \omega^j_k=\prod_{l=1}^{n_z} \int_m p(z_k^{[l]}|\{b_{k}^m\}^j) \{b_{k-1}^m\}^j dm.
\end{equation}

The brief implementation of RBPF-based confidence-rich localization and mapping is shown in Algorithm 1.
\begin{algorithm}[ht] 
	\caption{RBPF-CLAM(~)} 
	\begin{algorithmic}[1]
		\Require {Previous particles set $S_{k-1}$, previous expected map occupancy $\hat m_{k-1}$, the most recent sensor observation $z_k$ and odometry measurement $u_{k-1}$}
		\State $S_k \leftarrow \{\}$
		\For {$j$th particle in $S_{k-1}$}
		\State {$\{x_{k-1}^j,\omega_{k-1}^j\} \leftarrow s_{k-1}^j$}
		\State $//~\textit{Propagate the pose}$
		\State $x_k \leftarrow x_{k-1} \oplus u_{k-1}$
        \State $//~\textit{Update the weights}$
        \ForAll{beams in $z_k$}
        \State //~\textit{Compute Measurement Likelihood} \cite{yang2021crmi}
        \State $p_z$ = MeasurementLikelihood$(\hat m_{k-1},x_k,z_k)$
        \State $\omega_k^j \leftarrow \omega_k^j \cdot p_z$ 
        \EndFor
        \State {$//~\textit{Confidence-rich mapping}$ \cite{yang2021crmi}}
        \State $b_k^m, \hat m_k$ = ConfidenceRichMapping$(\hat m_{k-1},x_k,z_k)$
        \State $//~\textit{Update the particle set}$
        \State $S_k \leftarrow S_k \bigcup s_{k-1}^j$
        \EndFor 
        \State $N_{eff}= 1/\sum_1^{n_p}(\bar \omega^i_k)^2$
        \State $//~\textit{Resampling the particles}$
        \If {$N_{eff}<n_p/2$}  {$S_k \leftarrow$ Resampling($S_k$)}
        \EndIf \\
		\Return $S_k$
	\end{algorithmic} 
	\label{alg:rbpf}
\end{algorithm}

\section{Confidence-rich Mutual Information with Pose Uncertainty}
In this section, we study further the CRMI-based exploration considering pose uncertainty and present a solution to measure the expected uncertainty of the forthcoming pose.

Traditional information-based exploration methods mainly focus on minimizing the map uncertainty by choosing the optimal policies \cite{julian2014mutual,nelson2015information}. 
Several previous works have introduced the entropy about the posterior trajectory combined with map entropy to evaluate the candidate actions in active SLAM \cite{bourgault2002information,valencia2012active,vallve2015active}. Actually, it is difficult to compute this entropy because the current pose depends on the previous one under the Markov assumption \cite{stachniss2005information}. Hence, one has to approximate this entropy, such as by averaging the trajectory entropy without considering the correlations between poses \cite{valencia2012active}, or facilitating the computation under the multivariate Gaussian assumption over poses \cite{vallve2015active, popovic2020informative}. 

For a particle filter approximating the pose belief, a straightforward method to estimate the pose uncertainty is to use all normalized weights in a discretized way \cite{fischer2020information}:
\begin{equation}
    \label{eq:hx1}
    H(bel(x))
    =-\int bel(x)\log{bel(x)}d x\approx -\sum_{j=1}^{n_p} \omega_k^j \log{\omega_k^j}.
\end{equation}

Though this measure is a basic value to reflect the uncertainty reduction when updating particles and corresponding weights \cite{boers2010particle,radmard2018resolving}, this inaccurate approximation only relies on the weights to simulate the probability densities, and the essential information such as poses are ignored. Moreover, the resampled particles own the same weight, which can not reflect the distribution even worse.

Instead, we can express the entropy in a Bayesian manner using the state transition and measurement models. Applying the Bayes' rule to Eq.~\eqref{eq:hx1} yields the conditional entropy:
\begin{align}
    \label{eq:hx2}
    &H(bel_k(x))= -\int bel_k(x)\log{bel_k(x)}dx_{1:k} \nonumber\\
    &=-\int bel_k(x)\log{\frac{\overline{bel}_k(x)p(z_k|x_{1:k})}{p(z_k|z_{1:k-1})}}dx_{1:k} \nonumber\\
    &=-\int bel_k(x)\log{(\overline{bel}_k(x)p(z_k|x_{1:k}))}dx_{1:k} \nonumber\\
    &+ \log(p(z_k|z_{1:k-1})),
\end{align}
where the left term can be approximated by the particles:
\begin{align}
    \label{eq:left}
   \sum_{j=1}^{n_p}\log\left( p(z_k|s_k^j)(\sum_{j'=1}^{n_p}p(s_{k}^{j'}|s_{k-1}^{j'})\omega_{k-1}^{j'}) \right)\omega_{k}^j
\end{align}
and the right term can also be rewritten via the approximation of weighted particles:
\begin{align}
    \label{eq:right}
    \log(p(z_k|z_{1:k-1})) &= \log(\int p(z_k|x_{1:k})p(x_{1:k}|z_{1:k-1})dx_{1:k}) \nonumber\\ 
    &\approx \log(\sum_{j=1}^{n_p}p(z_k|s_k^j)\omega_k^j).
\end{align}
Note that $p(z_k|s_k)=p(z_k|x_{1:k})=p(z_k|x_k)$ holds in the above formulations because of the conditional independence assumption of observations.

Therefore, the information gain over path can be approximated and computed recursively as follows:
\begin{equation}
    I_{p} = H(bel_{k-1}(x)) - H(bel_k(x)).
\end{equation}
Here, we can compute the pose uncertainty efficiently based on the RBPF-CLAM using Eq.~\eqref{eq:hx2}, \eqref{eq:left} and \eqref{eq:right}. 

Consequently, the new information function UCRMI for an exploration task can be constructed by a linear combination of trajectory entropy gain and CRMI:
\begin{equation}
    I_c=\alpha I_m+(1-\alpha)I_{p},
    \label{eq:ucrmi}
\end{equation}
where $\alpha\in [0,1]$ is the constant for balancing mapping and localization and it depends on a certain task. The algorithm implementation for UCRMI is omitted here for brevity.

\section{Evaluation and Discussions}
In this section, we conduct comparative simulations and experiments to evaluate the proposed methods.
The beam-based mixture sensor model for CRM mapping are referred to \cite{thrun2005probabilistic}, where $z_{\text{hit}} = 0.7, z_{\text{short}}=0.1, z_{\text{max}}=0.1, z_{\text{rand}}=0.1$, $\lambda_{\text{short}} = 0.2$~m. The numerical integration resolution for map belief update is $\lambda_m=10$. 
All simulations and experiments are conducted in MATLAB 2020b using a 3.6GHz Intel i3-9100F CPU and 16G RAM on a desktop PC.

\subsection{The RBPF-CLAM vs. RBPF-OGM Results}
In this dataset simulation, we aim to study the performance of the RBPF-CLAM method in an unstructured and cluttered indoor environment. We use one cluttered part of Deutsches Museum dataset \cite{hess2016real} containing noisy laser range data collected by a horizontal 2D LIDAR and odometry data by an inertial measurement unit. This dataset contains 5522 scans and $n_z=1079$ beams per scan distributed uniformly among a field-of-view (FOV) of 270$^\circ$. We reduce the max range for sensing to 8~m and set a sensor noise covariance of $diag(0.01m, 0.01rad)^2$, i.e. we only use a minor part of the data to test the proposed method. 
The environmental grid map size is $90~\mathrm{m} \times 90 ~\mathrm{m}$ and the map resolution is 0.2 m. Note that we apply a high-precision particle filter containing over 1000 particles and use full data to generate a trajectory as the ground truth for comparison.

\begin{figure}
	    \centering
    	\subfigure[Estimated and ground truth trajectories in the resulting CRM]{
    		\begin{minipage}[t]{1\linewidth}
    			\centering
    			\includegraphics[width=0.8\linewidth]{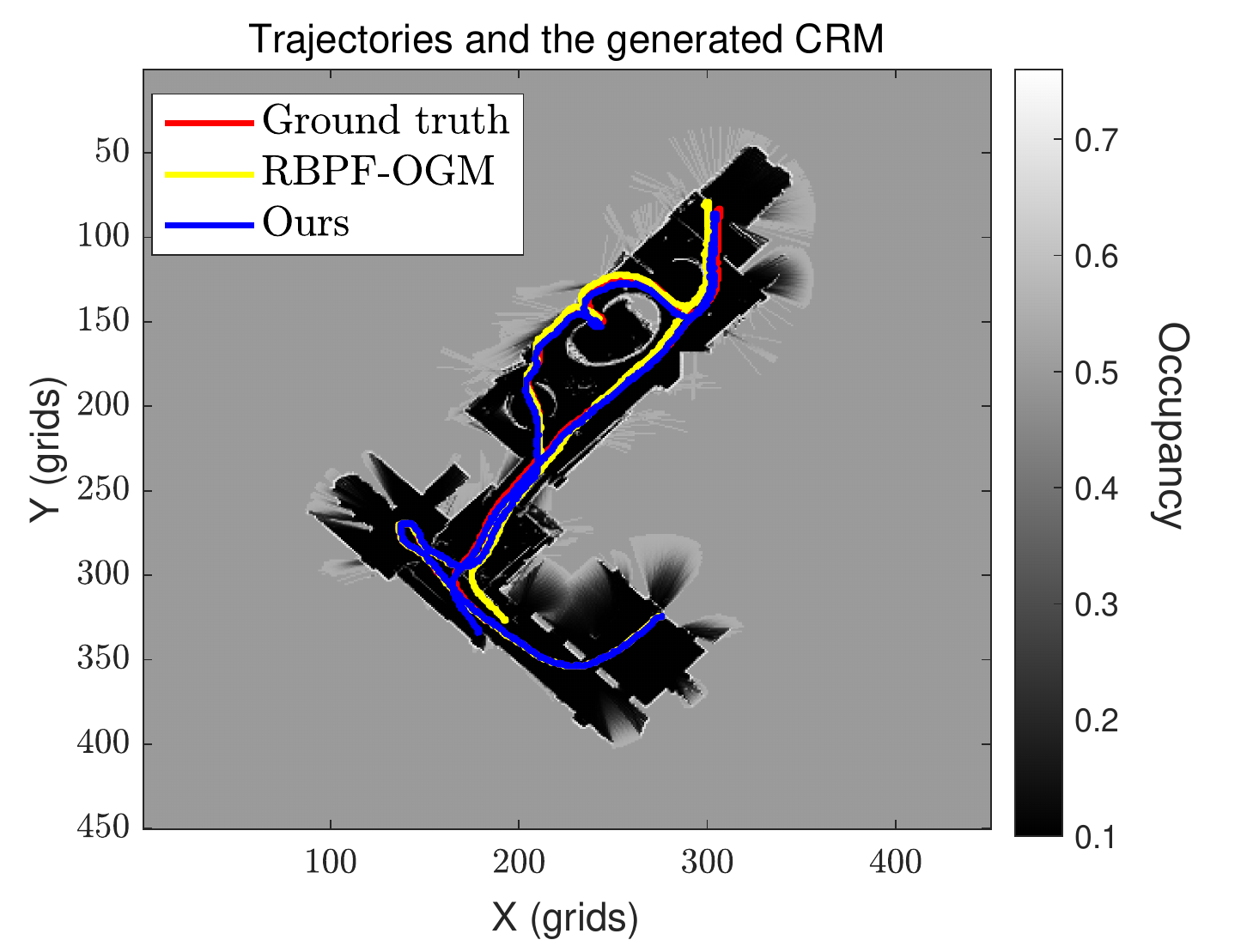}
    		\end{minipage}%
    	}%
    	\quad
    	\subfigure[Estimation error]{
    		\begin{minipage}[t]{1.0\linewidth}
    			\centering
    			\includegraphics[width=1.0\linewidth]{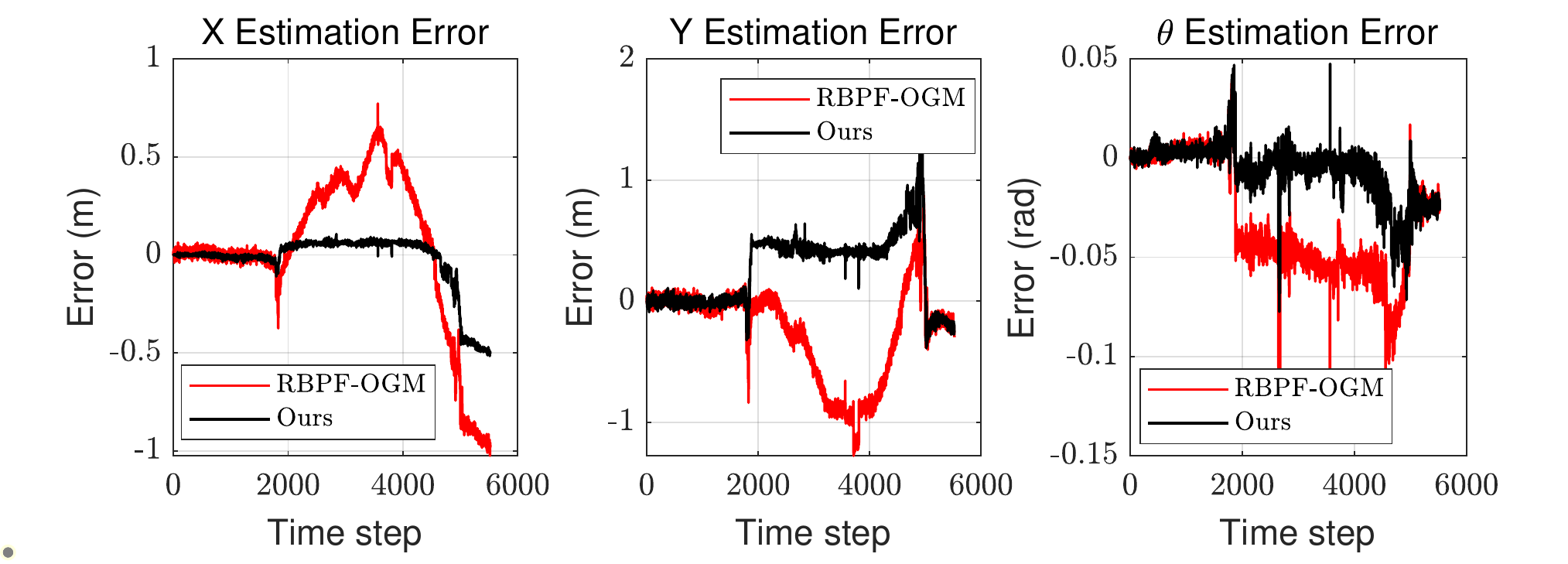}
    		\end{minipage}%
    	}%
		\caption{Illustration of the performance of RBPF-CLAM and RBPF-OGM in the Deutsche Museum dataset \cite{hess2016real}, a typical indoor complex environment. Our RBPF-CLAM has a relatively lower error than RBPF-OGM overall. The ground truth is obtained by a high-precision particle filter containing over 1000 particles.}
		\label{exp1}
	\end{figure}

The results are shown in Fig.~\ref{exp1}. We compare our RBPF-CLAM method with a typical RBPF SLAM approach \cite{grisetti2007improved} using OGMs representation. The numbers of particles are set to 100 for both methods. Fig.~\ref{exp1}(a) shows the trajectory estimation results and the estimation errors of $x,~y$ and $\theta$ are in Fig.~\ref{exp1}(b). 
Table~\ref{tab1} lists the mean absolute error (MAE) and average root mean squared error (RMSE) of 15 Monte Carlo experiments. 

Our RBPF-CLAM performs better than RBPF-OGM as expected, especially around the corners. This mainly attributes to the potential failure of scan matching in RBPF-OGM, for example, the significant estimation error around time steps 1800 and 4900. Instead, our proposed method computes the particles' weights using the closed-form measurement likelihood derived from posterior belief over map occupancy, which will improve the localization accuracy. 

\begin{table}[htbp]
	\caption{Localization errors: MAE and average pose RMSE}
	\begin{center}
		\begin{tabular}{c|c|c}
			\hline
			~ & RBPF-OGM& Ours\\ \hline
			 MAE $x$ (m) & 0.2780 $\pm$ 0.2766  & \textbf{0.0832 $\pm$ 0.1277}  \\ \hline
			 MAE $y$ (m) & 0.3108 $\pm$ 0.3357 & \textbf{0.3048 $\pm$ 0.2468} \\ \hline
			 MAE $\theta$ (rad) & 0.0344 $\pm$ 0.0255 & \textbf{0.01 $\pm$ 0.012} \\ \hline
			 Average RMSE (m) & 0.3191 $\pm$ 0.2824 & \textbf{0.2395 $\pm$ 0.1765} \\ \hline
		\end{tabular}
		\label{tab1}
	\end{center}
\end{table}

\subsection{Sampling-based Offline Informative Path Planning for Information Gathering Task}
\begin{figure}
	\centering
	\subfigure[CRMI-based IIG-tree \cite{yang2021crmi}]{
		\begin{minipage}[t]{0.5\linewidth}
			\centering
			\includegraphics[width=1.6in]{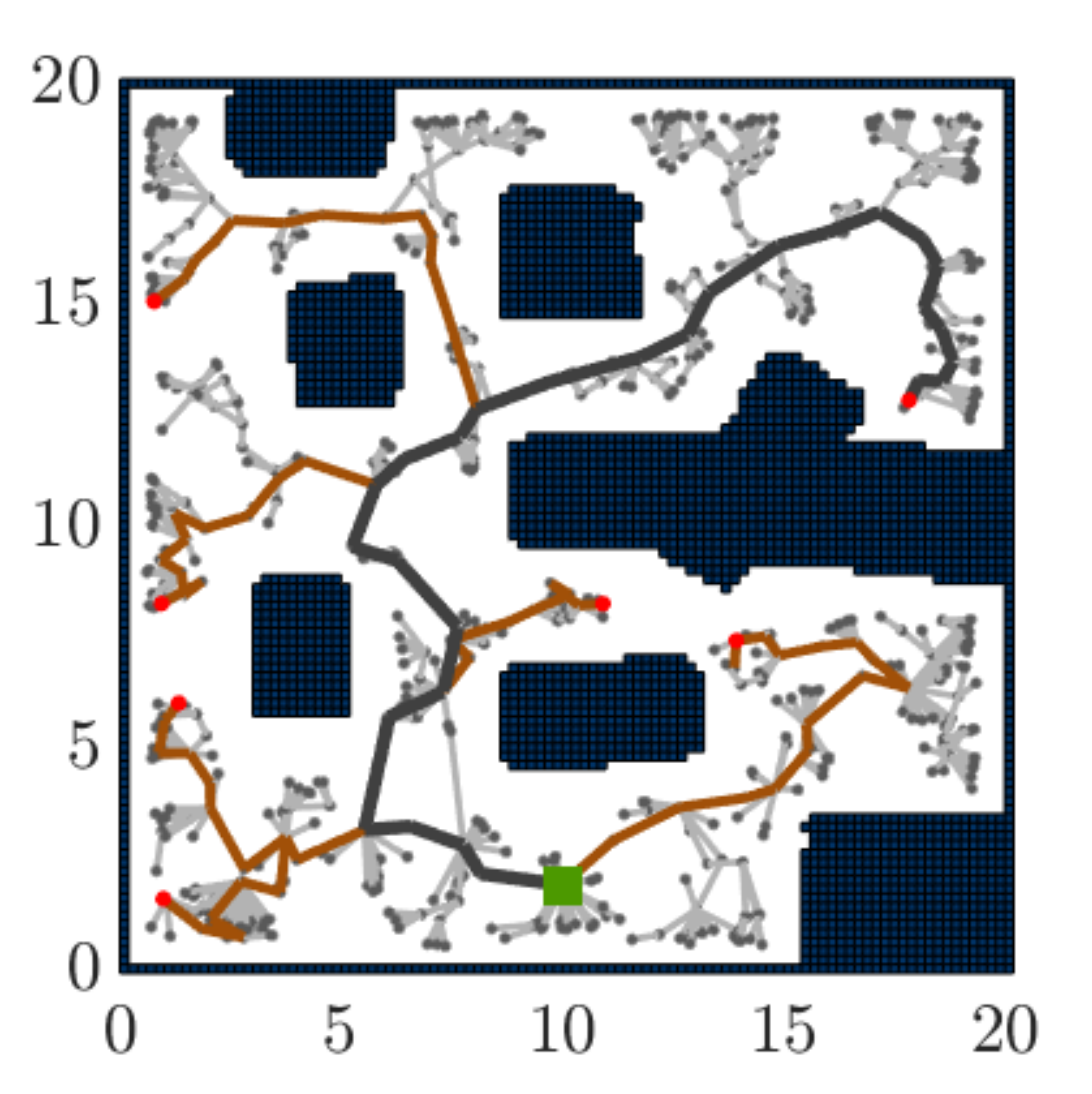}
		\end{minipage}%
	}%
	\subfigure[UCRMI-based IIG-tree]{
		\begin{minipage}[t]{0.5\linewidth}
			\centering
			\includegraphics[width=1.6in]{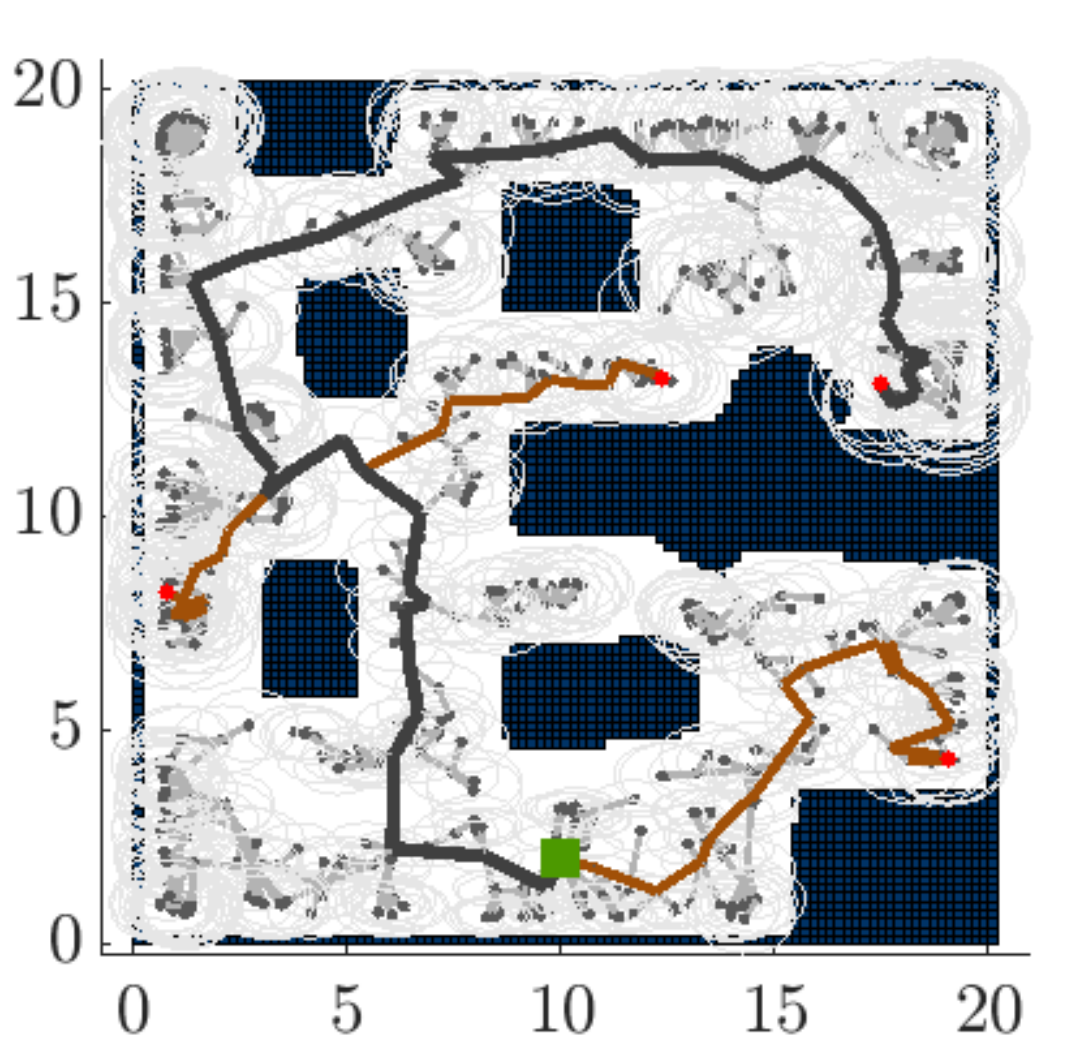}
		\end{minipage}%
	}%
	\quad
	\subfigure[OGMI-based IIG-tree \cite{julian2014mutual}]{
		\begin{minipage}[t]{0.5\linewidth}
			\centering
			\includegraphics[width=1.6in]{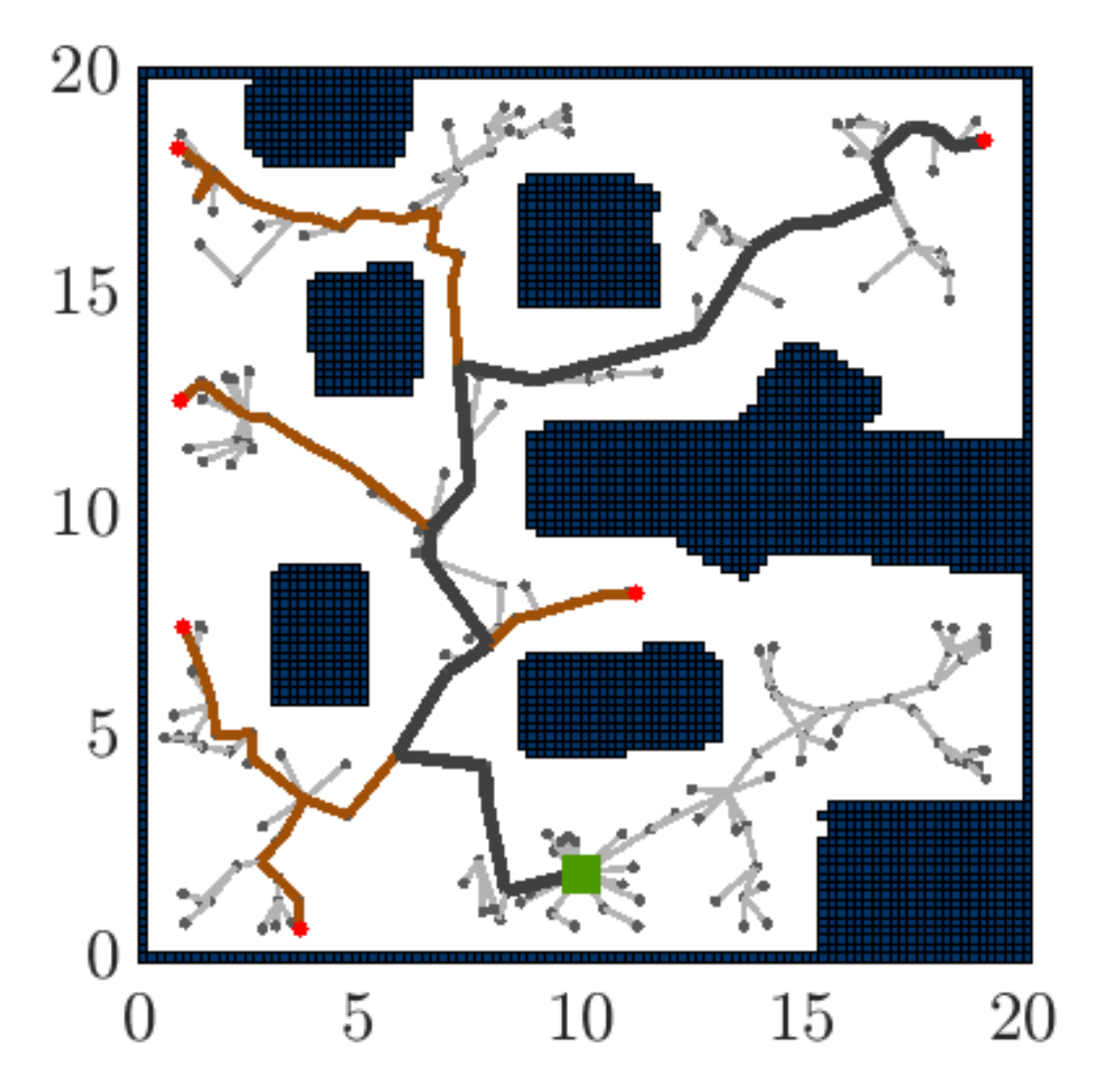}
		\end{minipage}%
	}%
	\subfigure[UGPVR-based IIG-tree \cite{ghaffari2019sampling}]{
		\begin{minipage}[t]{0.5\linewidth}
			\centering
			\includegraphics[width=1.6in]{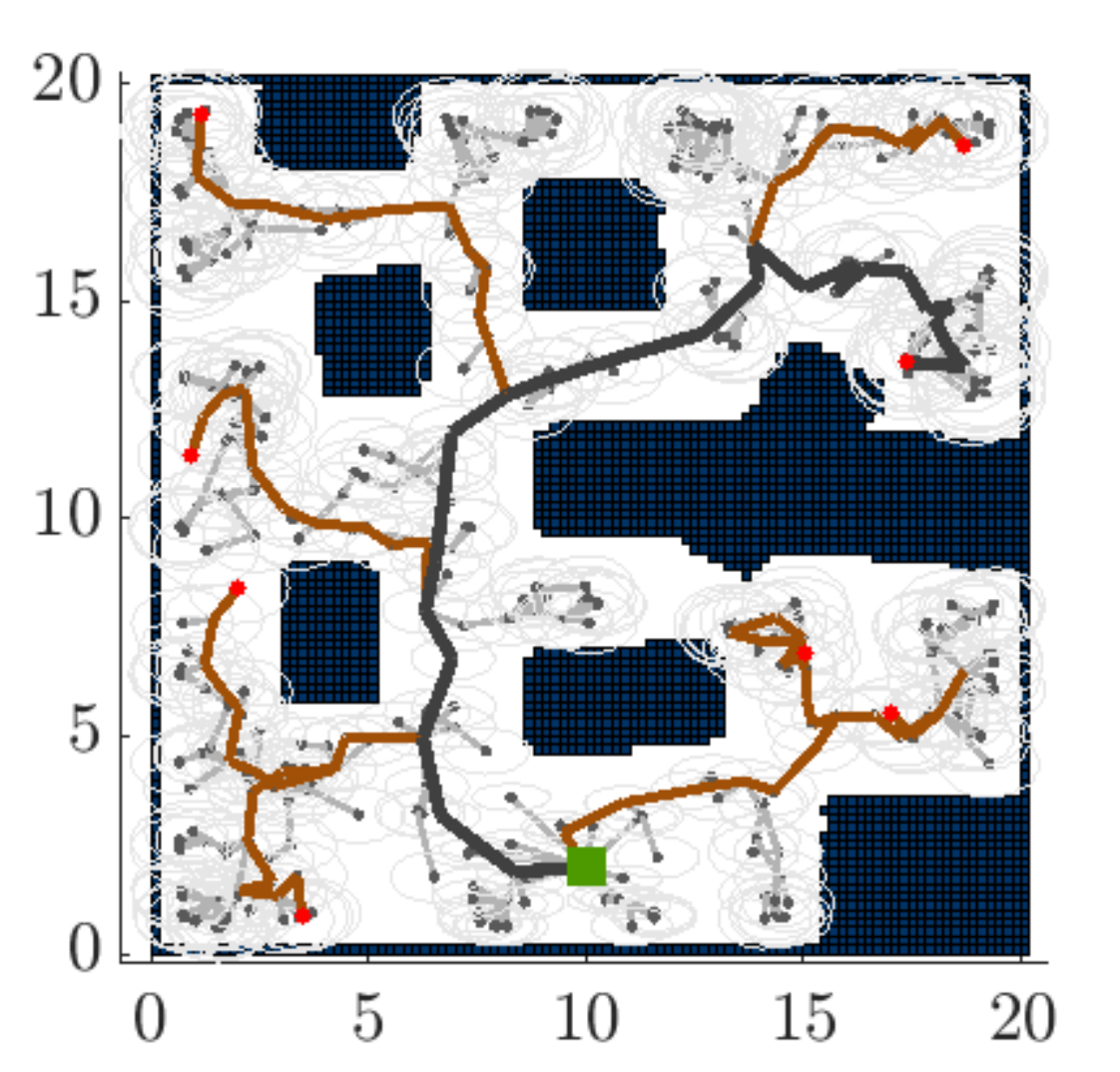}
		\end{minipage}%
	}%
	\centering
	\caption{The resulting IIG-graph using different information functions in one trial. These plots show respective IIG-trees and the most informative paths, the white ellipse denotes the pose uncertainty of each node. Note that the GPVR-based IIG-graph is omitted for brevity. Our UCRMI provides more thorough sampling and a more informative path than others. (\textbf{grey}: generated branches; \textbf{brown}: generated paths; \textbf{dark grey}: most informative path; all axis are in meters.)}
	\label{expl}
\end{figure}

\begin{figure}[ht]
	\centering
	\subfigure[Planning time]{
		\begin{minipage}[t]{0.5\linewidth}
			\centering
			\includegraphics[width=1.8in]{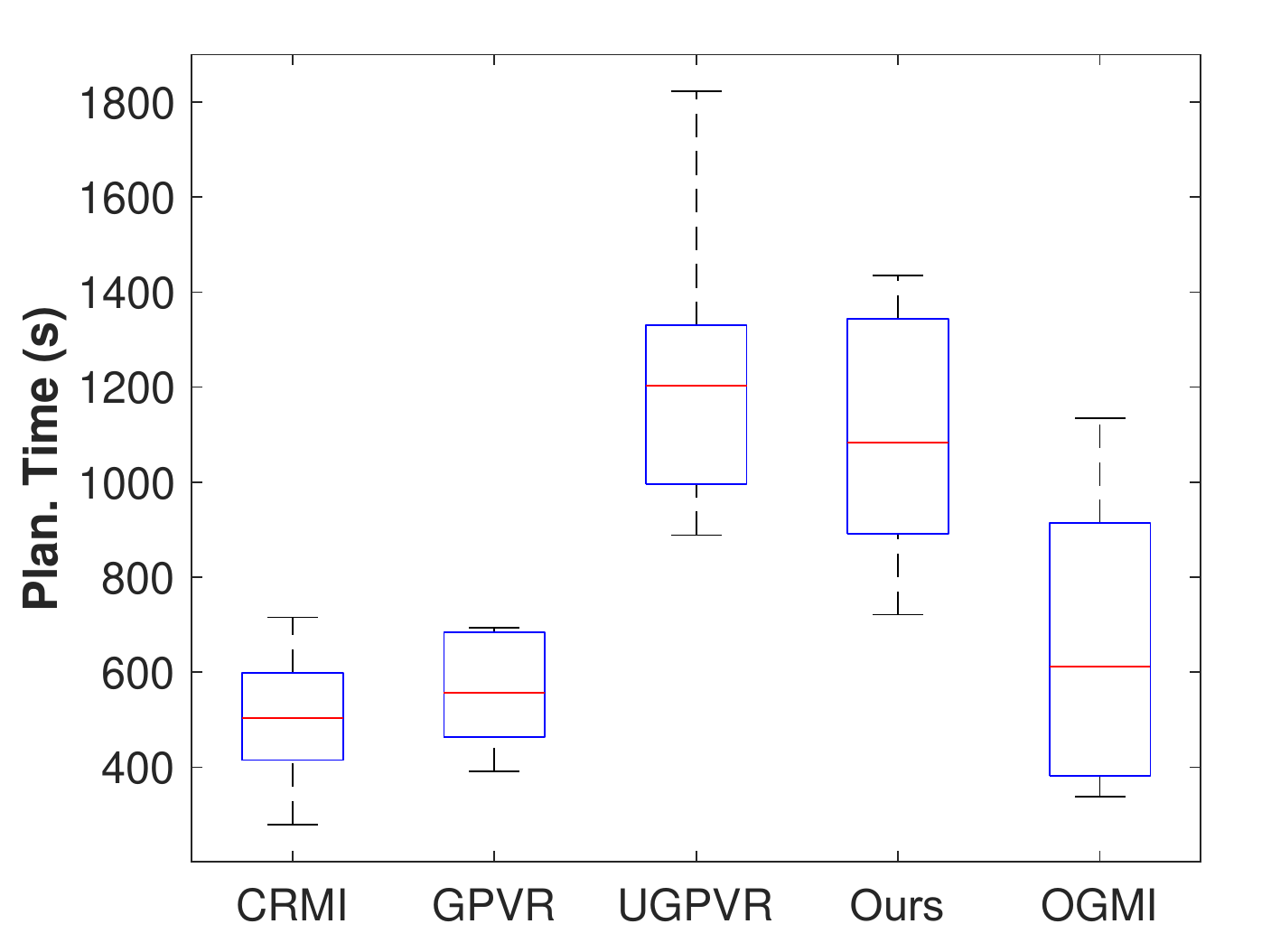}
		\end{minipage}%
	}%
	\subfigure[Total cost]{
		\begin{minipage}[t]{0.5\linewidth}
			\centering
			\includegraphics[width=1.8in]{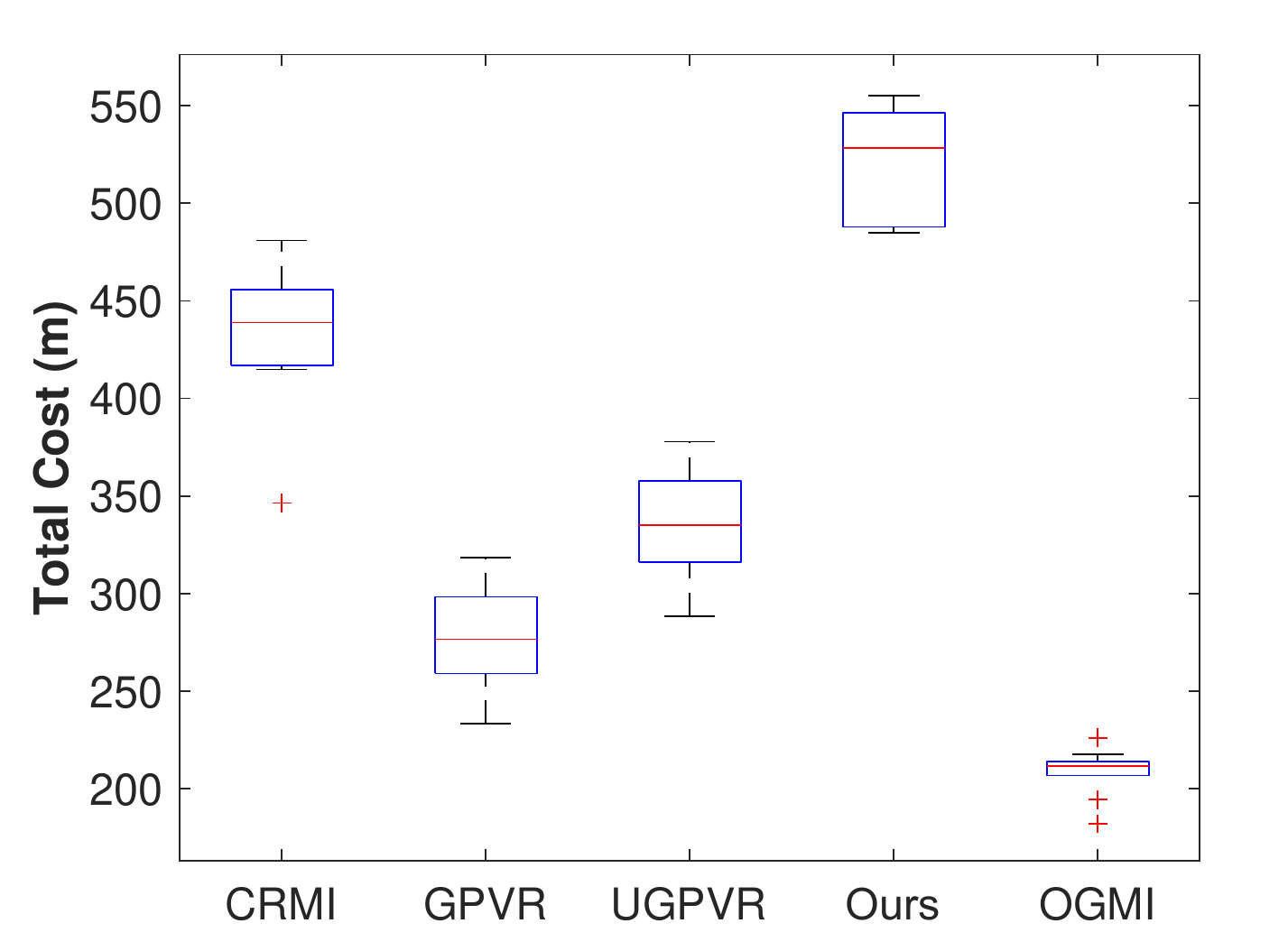}
		\end{minipage}
	}%
	\centering
	\caption{Box plot for IIG-graph using different information functions in 20 offline Monte Carlo experiments.}
	\label{box}
\end{figure}

\begin{figure}[ht]
	\centering
	\subfigure[Cumulative penalized $I_{RIC}$]{
		\begin{minipage}[t]{0.5\linewidth}
			\centering
			\includegraphics[width=1.6in]{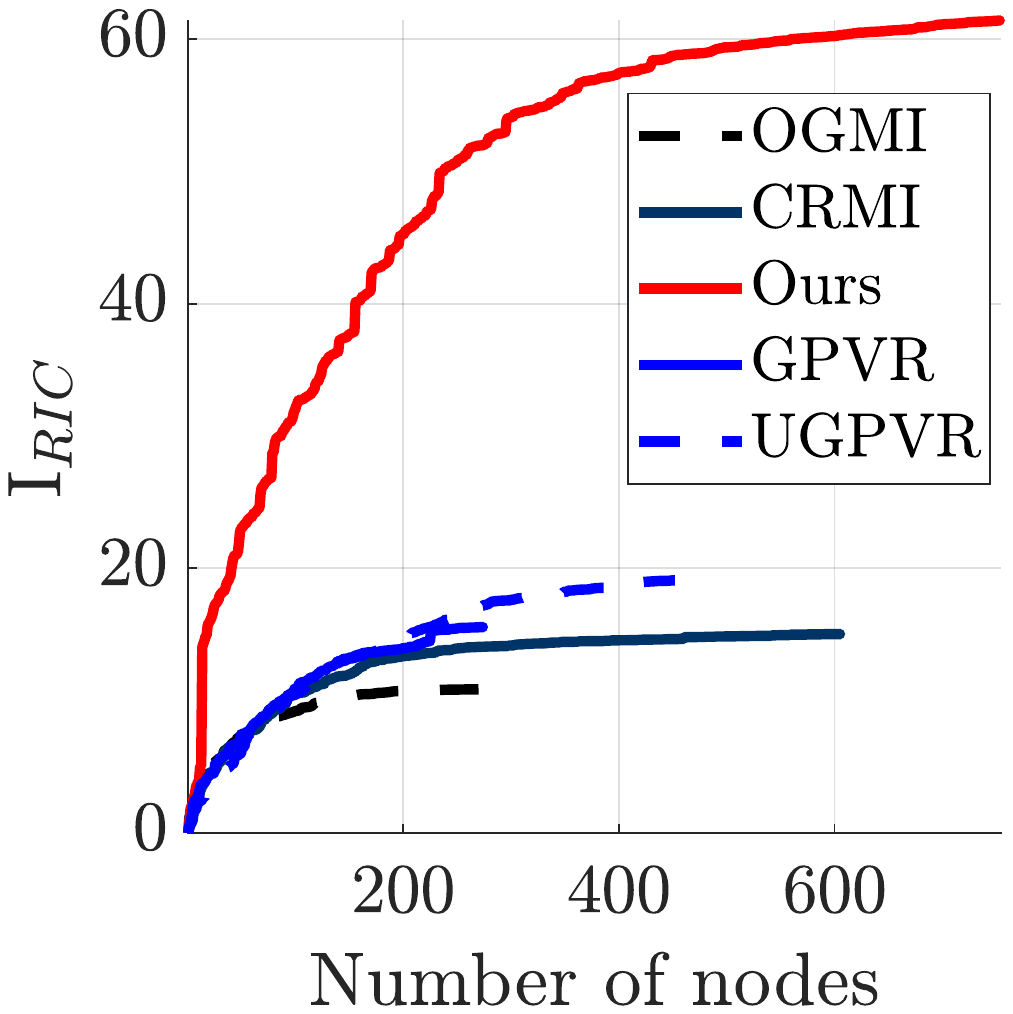}
		\end{minipage}%
	}%
	\subfigure[Evolution of information gain]{
		\begin{minipage}[t]{0.5\linewidth}
			\centering
			\includegraphics[width=1.6in]{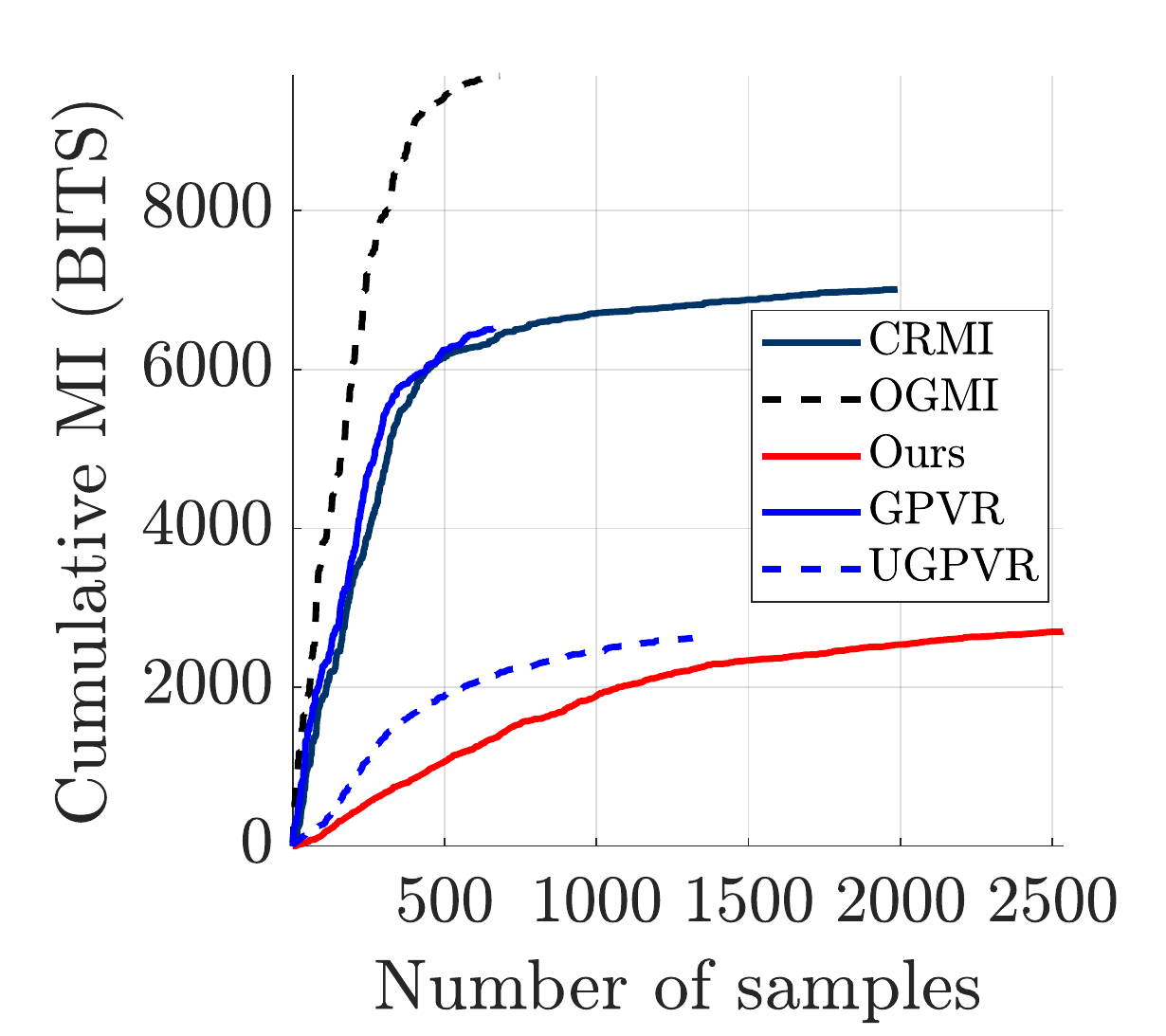}
		\end{minipage}%
	}%
	\centering
	\caption{Quantitative comparison of 5 different information functions in one trial. The results show the convergence of $I_{RIC}$ and the evolution of total information gain. Notice that our UCRMI owns the max average $I_{RIC}$ and more samples, showing more informative exploration.}
	\label{curve}
\end{figure}

In order to evaluate the exploration performance of our proposed information function UCRMI appropriately, we introduce the incrementally exploring information gathering (IIG) planner proposed by \cite{ghaffari2019sampling} as an offline evaluation platform. Built on the sampling-based informative motion planning methods\cite{hollinger2014sampling,binney2013optimizing}, IIG planner analyzed the information-theoretic convergence with a stopping criterion of the penalized $I_{RIC}$ for completing the exploration and information gathering tasks automatically, and find the most informative path given kinematics and budget constraints such as energy, consequently.

The relative information contribution (RIC) of a new node $x_{\text{new}}$ is defined by the information function values of itself and its neighboring node $x_{\text{near}}$, i.e. $RIC:= I_{\text{new}}/I_{\text{near}}-1$.
The resulting penalized RIC is defined by RIC and the number of samples $n_{\text{sample}}$ it takes to find the node $x_{\text{new}}$, i.e. $I_{RIC}:= RIC/n_{\text{sample}}$. This measures the average relative information gain/contribution of a new child node against its parent node in the IIG graph within a time span. The benefit of $I_{RIC}$ mainly lies in its property of non-dimensional and independence of information functions, which is quite suitable for comparing the information functions despite different map representations and computation algorithms. A higher $I_{RIC}$ value represents a denser IIG graph. More details about IIG please see \cite{ghaffari2019sampling}.

Here we use several representative information functions such as OGMI\cite{julian2014mutual}, CRMI\cite{yang2021crmi}, and GPVR/UGPVR\cite{ghaffari2019sampling} for comparison in an information gathering task when exploring a confined Cave map \cite{Radish}. The map size is $20~\mathrm{m} \times 20~\mathrm{m}$ and the map resolution is 0.2~m. 
The simulated laser scanner consists of $n_z=10$ beams with a FOV of 360$^\circ$. The sensing range limit is $z_{\text{MAX}}=5$~m, and the numerical integration resolution for CRMI calculation is $\lambda_z=0.1$ m. We set the cost budget as $10^5$ and the planner convergence threshold $\lambda_{RIC}$ as 0.005. In addition, the constant $\alpha$ for UCRMI is set to 0.5 for balance the exploration and localization. The robot evaluates the expected information gain of each candidate child node via executing forward simulation using virtual ray casting to the map, then generates several candidate informative paths to choose the one with maximum information gain. 

Fig.~\ref{expl} shows the IIG-graph using 4 different information functions (GPVR figure omitted) and the corresponding most informative paths. Fig.~\ref{box} shows the specific statistical results comparison in box plots. Fig.~\ref{curve} presents the curves' evolution as the the nodes and samples increase.
Note that the total information gain and total cost of all edges are defined over the whole explored area.

Since the information functions are derived from different maps, sensor models and calculation, the above-mentioned $I_{RIC}$ play a key role in this analysis. A farther node can contribute much more information gain than a closer node because of the higher pose uncertainty in the former one.
Compared with other metrics, OGMI/GPVR/CRMI drive the faster but rough exploration (see Fig.~\ref{expl} and Fig.~\ref{curve}) since they generate fewer samples and nodes (see the X-axis in Fig.~\ref{curve}).
UGPVR also has fewer samples/nodes but spends even more planning time than UCRMI. Instead, as shown in Fig.~\ref{box}(b) and Fig.~\ref{curve}(a), our UCRMI plans the longest informative path and owns most nodes/samples than others, it makes the best use of the given distance budget than others. The inherent principle mainly lies in the pose uncertainty reduction brought by the RBPF-CLAM. 
Fig.~\ref{curve}(b) also shows UCRMI has the lower total information gain than CRMI because of the same reason.

Meanwhile, the computation cost of UCRMI increases after incorporating the one of pose uncertainty to CRMI. Thus, UCRMI and CRMI curves have similar evolution trends in Fig.~\ref{curve}, but the former converges slower than CRMI evidently, similar to the case of UGPVR slower than GPVR. Compared with others, UCRMI also has a longer tail before the planner converges, as in Fig.~\ref{curve}.
However, UCRMI has the higher $I_{RIC}$ than others, as mentioned before, this implies the spanning tree of UCRMI is much denser than others and it will conduct a more meticulous exploration.

Essentially, the RBPF-CLAM can prevent an UCRMI-driven robot explores greedily and aggressively in unknown areas for safety concern. This is mainly attributed to the RBPF-CLAM can help UCRMI reduce the information gained from new measurements at the candidate position near the unexplored areas by the pose uncertainty reduction (c.f. Eq.~\eqref{eq:ucrmi}), and keeping the robot staying at places easier to localize. 

In short, our UCRMI provides a full state estimate of the map and robot pose in one integrated measure. It guides more prudent, fine, and safe exploration actions than other information functions in the information-gathering process of unknown environments. Simulation results also evidence this.


\section{Conclusion}
This paper mainly contributed a new method of the Monte-Carlo localization of CRMs and incorporating measurable and more accurate pose uncertainty into CRMI-based robotic exploration. Particularly, the localization and mapping of CRMs are combined in an RBPF framework, and the resulting accuracy has been improved by a new weighting algorithm based on the closed-form measurement likelihood derived from the map posterior distribution. The pose information gain is also approximated by the particle filter. Dataset simulation and experimental results show the desired localization performance and exploration results in information-gathering tasks of unknown environments.
Our next work will study the exploration-exploitation trade-off using different $\alpha$ values and online active informative planning.

\bibliographystyle{IEEEtran}
\bibliography{iros2022}

\end{document}